# Evaluation of Point Pattern Features for Anomaly Detection of Defect within Random Finite Set Framework

Ammar Mansoor Kamoona
School of Engineering
RMIT University
Victoria, Australia
Email: ammar.kamoona@rmit.edu.au

Amirali Khodadadian Gostar
School of Engineering
RMIT University
Victoria, Australia
Email: amirali.khodadadian@rmit.edu.au

Alireza Bab-Hadiashar
School of Engineering
RMIT University
Victoria, Australia
Email: alireza.bab-hadiashar@rmit.edu.au

Reza Hoseinnezhad
School of Engineering
RMIT University
Victoria, Australia
Email: reza.hoseinnezhad@rmit.edu.au

*Abstract*—Defect detection in the manufacturing industry is of utmost importance for product quality inspection. Recently, optical defect detection has been investigated as an *anomaly detection* using different deep learning methods. However, the recent works do not explore the use of point pattern features, such as SIFT for anomaly detection using the recently developed set-based methods. In this paper, we present an evaluation of different point pattern feature detectors and descriptors for defect detection application. The evaluation is performed within the random finite set framework. Handcrafted point pattern features, such as SIFT as well as deep features are used in this evaluation. Random finite set-based defect detection is compared with state-of-the-arts anomaly detection methods. The results show that using point pattern features, such as SIFT as data points for random finite set-based anomaly detection achieves the most consistent defect detection accuracy on the MVTec-AD dataset.

## I. Introduction

Automated visual inspection is a key part of the quality control process in many manufacturing applications and overpowers the limitation of human visual inspection. Manual visual inspection errors commonly include missing or incorrect identification of defects. Such errors can have a significant impact on product quality and may lead to unnecessary increase in production cost and overall waste. Therefore, an automated computer vision-based visual inspection is the key to productivity [1]. Different data-driven methods based on deep learning have been proposed for visual inspection in different areas, such as manufacturing [2]–[5], civil engineering [6]–[8], transportation [9], [10] and computing systems [11], [12].

Earlier, a number of automated solution has been developed using different image processing algorithms such as Haar filler for tile surface inspection [13], local order binary pattern for fabric defect detection [14], and scale-invariant keypoint features for PCB inspection [15]. Furthermore, a combination of image processing and machine learning algorithms has shown a satisfactory performance. An example of this is railway inspection using the histogram of oriented gradients (HOG) with support vector machine (SVM) [16]. Fabric inspection using adaptive boosting with HOG features and SVM was also proposed in [17]. However, a major drawback of image processing techniques is their focus on the use of implicit engineered features which can be very challenging when describing complex cases.

Recently, the success of deep learning methods in computer vision [18], [19] using convolution neural networks has paved the way to apply deep learning for visual inspection. These methods use the data representation learning to perform different tasks where the goal is to transform complex data to abstract representations known as features. Wang *et al.* [4] highlighted the power of deep learning for smart manufacturing and how this changes the future industry trends. Different deep learning tools have been proposed to deal with different surface defects [20], [21]. Hui *et al.* [22] proposed a LEDNet network for defect detection and classification on LED chips. Cha *et al.* [6] proposed structural damage detection using Faster R-CNN to detect five different defects.

The use of deep learning methods, such as convolutional neural networks (CNNs) is limited by the availability of the training samples which gives rise to two problems: class imbalance between the normal and defected samples and the difficulty of data annotation. Collecting a large number of unlabeled images (normal and with defect) is simple but labeling these samples is expensive and requires a trained inspector. These two problems are well-know and subject of continuing research [23]–[25]. Therefore, treating defect detection as an anomaly detection problem is the ideal solution for automated visual defect detection due to the lack of defective samples and their uncertainty. Recently, Bergmann *et al.* [26] proposed an evaluation of different deep anomaly detection methods for defect detection. They also presented a new dataset (MVTec) which includes a variety of different objects and textures with different types of defects. However, the aforementioned work does not explore the use of point pattern features for defect detection within Random finite set framework.

Vo et al. [27] proposed to model the point pattern features (e.g SIFT) using point processes via multiple instance learning for anomaly detection. The proposed approach is based on treating each point pattern feature as a sample of random finite set (RFS). In addition, a likelihood function based on the RFS density assumptions (e.g Possion IDD clusters) was proposed. The proposed approach allows to incorporate both the number of extracted features (cardinality) information and feature density information into the anomaly detection solution. In this framework, the cardinality of features contributes to the RFS density.

In this paper, we provide an evaluation of different point pattern features for defect detection within RFS framework. The main goal of this study to explore the advantage of RFS-based anomaly detection for defect detection and to show how this approach can generate competitive results compared with current state-of-the-art deep learning approaches. In this evaluation, we have examined different state-of-the-art point pattern features that have been used for matching, as well as the well-known SIFT keypoint detector and descriptor.

## II. RANDOM FINITE SET-BASED DEFECT DETECTION

The random finite set-based anomaly detection method models the point pattern features as an RFS. The rationale is that both the feature elements $x_i$ ($i = 1, \ldots, n$) in the set $X$ and the number of features, $|X| = n$, vary randomly. Accordingly, each measured set of point features $X = \{x_i\}_{i=1}^{|X|}$ is treated as an RFS. The RFS density with respect to some measure $U$ is given by [28]:

$$p(X) = p_c(|X|)(|X|)! U^{|X|} p(x_1, \ldots, x_{|X|}), \quad (1)$$

where $p_c(n) = P(|X| = n)$ is the cardinality distribution, $U$ is the unit hyper-volume, and $p(x_1, \ldots, x_n)$ is a symmetric joint feature density for the given cardinality $|X| = n$ [29].

In practice, different assumptions about the mathematical form of the RFS density $p(X)$ can be considered. Examples include, Poisson RFS [30], Beta RFS [31], the Bernoulli RFS [28], the multi-Bernoulli RFS [32], the IDD-cluster RFS [33], the labeled multi-Bernoulli RFS [34] and finally the generlized labeled multi-Bernoulli RFS [35]. The aforementioned densities have been used for multi-object tracking in which they treat multi-object entity as a random finite set. An independent identical distributed (IDD) cluster RFS density as follows:

$$p(X) = p_c(|X|)(|X|)! U^{|X|} [p(\cdot)]^X, \quad (2)$$

where $p(\cdot)$ is the feature density and $[p(\cdot)]^X$ is the finite set exponential. With IDD cluster RFS, we need to make two different assumptions about the mathematical form of the $p_c$ and $p(\cdot)$. If we assume the cardinality distribution follows Poisson distribution. Then, the IDD-cluster RFS turns into Poisson RFS given by:

$$p(X) = \rho^{|X|} \exp(-\rho) [Up(\cdot)]^X, \quad (3)$$

where $\rho$ is the non-negative Poisson intensity.

## III. EVALUATION METHODOLOGY

In this section, we discuss the use of random finite set-based anomaly detection for defect detection. The general proposed approach of the RFS defect detection is shown in Figure 1. The proposed approach consists of two steps. The first step is a point pattern feature extractor and descriptor and the second part is a feature modeling and defect detection using RFS framework.

### A. Point pattern feature extractor and descriptor

The term *point pattern feature* refers to any feature extraction pipeline that returns a set of keypoints (interest points) rather than vector based feature. These keypoints are 2D locations in an image which should be stable and repeatable against different lighting conditions and viewpoints. Point pattern features have been used in different computer vision tasks, such as Simultaneous Localization and Mapping (SLAM), camera calibration, Structure-from-motion (SfM) and image matching [36]. Generally, most of the local feature detection methods are in the form of a set such as in SIFT [15]; while global feature detection methods are commonly return features in vector format, such as in the Histogram of Oriented Gradients (HOG) [37]. In machine learning, the traditional approach is to convert the point pattern features into vector format by using different methods such Bag of visual world [38].

The most well-known handcrafted point pattern feature detection method that has been used in this evaluation is a Harris-Laplace point detector which uses Harris corner detector to detect keypoints which are scale invariant. Then, a descriptor such as SIFT [15] around these keypoints is calculated where the size of the area depends on the maximum scale of the Laplacian-of-Gaussians [39]. SIFT descriptor was proposed by Lowe [15] to describe the local shape around keypoints using the edge orientation histogram. SIFT descriptor has shown a good performance in different local feature localization and matching applications. Different color variants of SIFT have been proposed [40], such as Hue-SIFT, color-SIFT and opponent-SIFT, to address the illumination variations issues. Evaluation of different handcrafted keypoints detection and descriptors is performed in [41], [42].

Deep learning using convolution neural networks has shown to be superior to handcrafted features representation in different computer vision tasks, such as human pose estimation [43] and object detection algorithms [44] that require images as input. As a result, convolution neural network has been used for keypoint detection and descriptor in many applications [45]–[48].

In this evaluation, we use a different point pattern handcrafted and deep learned feature set with the Random finite set framework for feature learning and defect detection.

### B. Defect detection using RFS

Due to the lack of access to the defected samples, unsupervised anomaly detection is a preferred option for defect detection [26]. In this approach, only the normal samples (defect-free) are used in the training phase. Similarly, the RFS-based defect detection only uses the normal samples during training to maximize RFS set density in which the parameter of the model could be learned using either the maximum likelihood estimator (MLE) or expectation maximization (EM) [31].

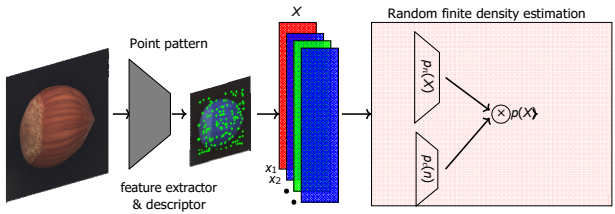

Fig. 1: The general proposed approach for RFS-based defect detection. The highlighted part is the only part that can be trained.

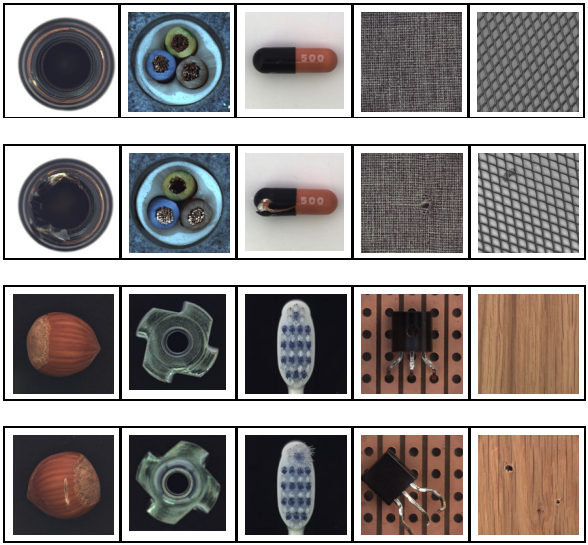

Fig. 2: Samples of MVTec-AD dataset, defect free (first row) and defected samples (second row).

## IV. EXPERIMENTS

### A. Dataset

**MVTec AD dataset**: MVTecAD [26] is a comprehensive and challenging real-world industrial image dataset that is developed for defect detection. The dataset has an extensive collection of texture and object images. It has 5354 high-resolution color images of a variety of objects and textures. In MVTech AD, there are 15 different categories (ten objects and five textures). Each object has only normal samples for training and normal and defect samples for testing. There are 70 different types of defects, such as scratches, dents, contamination, and various structural deformations. Figure 2 shows examples of these samples. The first row shows the defect-free samples, while the second row shows defect samples.

### B. feature extraction

In order to extract sparse local features from the defect-free samples, a different point pattern features extraction methods have been used in this evaluation. The focus of this paper is on the current state-of-the-art deep learned features. Thus, the following deep learning point pattern feature detection and descriptor has been used. **LF-net** [46]: is an unsupervised learned network that use detect-then-describe strategy in one end-to-end network. the network output a descriptor with dimension 255-D. **D2-net** [47]: this network provides joint learning for detection and description. During the inference, the network provide a multi-scale option, in which we call this network as (D2-net2). **r2d2** [48]: this network provides joint learning of detection and description. This network learns both keypoint repeatability and confidence from the training set. r2d2 network is a self-supervision trained network using synthetic and real images.

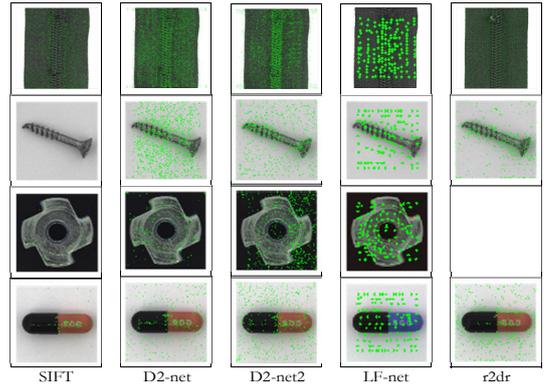

Fig. 3: Point pattern features (keypoints) of different feature extraction methods.

Finally, we have used the most well-known handcrafted feature detection and description which is **SIFT** [15] and Harris-Laplace point detector because they have shown good performance for category recognition. Figure **??** shows the keypoints of different feature extraction.

### C. Experimental Results

The proposed RFS-based defect detection using different point pattern feature detection and description is evaluated on MVTec AD dataset. For the sake of comparison, we have compared the proposed approach with the state-of-the-art deep anomaly detection given in [26] which are as follows: **AE(SSIM)**: deep auto-encoder using SSIM as loss function. **AE(L2)**: deep auto-encode using pixel wise $L_2$ loss function. **AnoGAN**: anomaly detection using Generative adversarial networks. **CNN**: convolutional neural network feature dictionary [49]. **Texture inspection**: Gaussian mixture model for texture inspection [50]. **variation model**: using varitional model [51] of GMM for non-texture images is used by providing prior alignment of the object countours. The implementation details and setups of these methods can be found in [26]. As an **Evaluation metric**, we have used the same approach used in [26] in reporting the defect detection accuracy. We have calculated the accuracy of correctly classifying images for normal and defect samples.

Table I shows the ratio of correctly classified (normal and defect) samples for each object and the mean of these ratios. We rank the mean of each object (the lower is better) of these methods and the final rank of the average (Avg.) rank is shown in the last row. It is clear that the RFS-based defect detection using (SIFT) has the best performance (ranks first) followed with auto-encoder-based methods (rank second and third). RFS (LF-net) and RFS (r2d2) shows better performance compared to AnoGan and CNN-dictionary methods. We can see that RFS-based defect detection shows a promising performance compared with the state-of-the-art methods for object-based defect detection.

TABLE I: Performance results of Poisson RFS-based defect detection using different feature extractions methods on MVTec-AD dataset. The ratio of corrected classified normal samples (top row) and ratio of corrected classified abnormal samples (bottom row) are given. The best means are in **bold**.

| Category | RFS (SIFT) | RFS (Harris) | RFS (D2-net) | RFS (D2-net2) | RFS (LF-net) | RFS (r2d2) | AE (SSIM) | AE (L2) | AnoGAN | CNN | Variation Model |
|---|---|---|---|---|---|---|---|---|---|---|---|
| Zipper | 0.75 | 0.75 | 0.59 | 0.78 | 0.00 | 0.00 | 1.00 | 0.97 | 0.78 | 0.78 | - |
|  | 0.88 | 0.55 | 0.77 | 0.84 | 1.00 | 1.00 | 0.60 | 0.63 | 0.40 | 0.29 | - |
| Mean | **0.81** | 0.65 | 0.68 | **0.81** | 0.50 | 0.50 | 0.80 | 0.80 | 0.59 | 0.54 | - |
| Rank | 1 | 4 | 3 | 1 | 7 | 7 | 2 | 2 | 5 | 6 |  |
| Transistor | 0.88 | 0.81 | 0.96 | 0.96 | 0.93 | 0.91 | 1.00 | 0.97 | 0.98 | 1.00 | - |
|  | 0.52 | 0.21 | 0.20 | 0.20 | 0.07 | 0.10 | 0.03 | 0.45 | 0.35 | 0.15 | - |
| Mean | 0.70 | 0.50 | 0.58 | 0.58 | 0.50 | 0.50 | 0.51 | **0.71** | 0.66 | 0.57 | - |
| Rank | 2 | 7 | 4 | 4 | 7 | 7 | 6 | 1 | 3 | 5 |  |
| Toothbrush | 0.75 | 0.91 | 1.00 | 0.91 | 0.91 | 0.75 | 0.75 | 1.00 | 1.00 | 1.00 | 1.00 |
|  | 0.60 | 0.30 | 0.06 | 0.03 | 0.50 | 0.06 | 0.73 | 0.97 | 0.13 | 0.03 | 0.60 |
| Mean | 0.67 | 0.60 | 0.53 | 0.47 | 0.70 | 0.40 | 0.74 | **0.98** | 0.56 | 0.51 | 0.80 |
| Rank | 5 | 6 | 8 | 10 | 4 | 11 | 3 | 1 | 7 | 9 | 2 |
| Screw | 0.63 | 0.78 | 0.65 | 0.70 | 0.56 | 1.00 | 0.95 | 0.98 | 0.41 | 0.73 | 1.00 |
|  | 0.86 | 0.73 | 0.58 | 0.57 | 0.70 | 0.99 | 0.06 | 0.39 | 0.28 | 0.13 | 0.10 |
| Mean | 0.74 | 0.75 | 0.62 | 0.63 | 0.63 | **0.99** | 0.50 | 0.68 | 0.34 | 0.43 | 0.55 |
| Rank | 3 | 2 | 6 | 5 | 5 | 1 | 8 | 4 | 10 | 9 | 7 |
| Metal nut | 0.68 | 1 | 0.5 | 0.68 | 0.77 | - | 1.00 | 0.68 | 0.86 | 0.55 | 0.32 |
|  | 0.79 | 0.30 | 0.56 | 0.58 | 0.52 | - | 0.08 | 0.77 | 0.13 | 0.74 | 0.83 |
| Mean | **0.735** | 0.65 | 0.53 | 0.63 | 0.64 | - | 0.54 | 0.72 | 0.49 | 0.37 | 0.57 |
| Rank | 1 | 3 | 8 | 5 | 4 | - | 7 | 2 | 9 | 10 | 6 |
| Pill | 0.84 | 0.84 | 0.76 | 0.69 | 0.96 | 0.80 | 0.92 | 1.00 | 1.00 | 0.85 | 1.00 |
|  | 0.50 | 0.23 | 0.27 | 0.38 | 0.17 | 0.72 | 0.28 | 0.23 | 0.24 | 0.06 | 0.13 |
| Mean | 0.67 | 0.54 | 0.52 | 0.53 | 0.56 | **0.76** | 0.60 | 0.61 | 0.62 | 0.45 | 0.56 |
| Rank | 2 | 7 | 9 | 8 | 6 | 1 | 5 | 4 | 3 | 10 | 6 |
| Bottle | 1.00 | 1 | 0.95 | 0.95 | 0.90 | 0.75 | 0.85 | 0.70 | 0.95 | 1.00 | 1.00 |
|  | 0.66 | 0.01 | 0.87 | 0.44 | 0.85 | 0.65 | 0.90 | 0.89 | 0.43 | 0.06 | 0.13 |
| Mean | 0.83 | 0.50 | **0.91** | 0.69 | 0.87 | 0.70 | 0.87 | 0.79 | 0.69 | 0.53 | 0.56 |
| Rank | 3 | 9 | 1 | 6 | 2 | 5 | 2 | 4 | 6 | 8 | 7 |
| Cable | 0.89 | 0.24 | 0.96 | 0.87 | 0.58 | 0.10 | 0.74 | 0.93 | 0.98 | 0.97 | - |
|  | 0.23 | 0.86 | 0.05 | 0.33 | 0.59 | 0.94 | 0.48 | 0.18 | 0.07 | 0.24 | - |
| Mean | 0.56 | 0.55 | 0.50 | 0.60 | 0.59 | 0.52 | **0.61** | 0.55 | 0.52 | 0.60 | - |
| Rank | 4 | 5 | 7 | 2 | 3 | 5 | 1 | 5 | 6 | 2 |  |
| Capsule | 0.78 | 0.73 | 1.00 | 0.82 | 0.26 | 0.78 | 0.78 | 1.00 | 0.96 | 0.78 | 1.00 |
|  | 0.48 | 0.45 | 0.08 | 0.46 | 0.84 | 0.67 | 0.43 | 0.24 | 0.20 | 0.03 | 0.03 |
| Mean | 0.63 | 0.59 | 0.54 | 0.64 | 0.55 | **0.73** | 0.60 | 0.62 | 0.58 | 0.40 | 0.51 |
| Rank | 3 | 6 | 9 | 2 | 8 | 1 | 5 | 4 | 7 | 11 | 10 |
| Hazelnut | 0.95 | 0.95 | 0.37 | 0.00 | 0.95 | 0.77 | 1.00 | 0.93 | 0.83 | 0.90 | - |
|  | 0.82 | 0.08 | 0.75 | 1.00 | 0.42 | 0.37 | 0.07 | 0.84 | 0.16 | 0.07 | - |
| Mean | **0.88** | 0.51 | 0.56 | 0.50 | 0.68 | 0.57 | 0.53 | **0.88** | 0.49 | 0.48 | - |
| Rank | 1 | 6 | 4 | 7 | 2 | 3 | 5 | 1 | 8 | 9 | - |
| Avg.Mean |  |  | 0.59 | 0.60 |  |  | 0.63 | 0.73 | 0.55 | 0.46 |  |
| Avg.Rank | 2.5 | 5.5 | 5.9 | 5 | 4.8 | 4.5 | 4.4 | 2.8 | 6.4 | 7.9 | 6.3 |
| Rank | **1** | 7 | 8 | 6 | 5 | 4 | 3 | 2 | 10 | 11 | 9 |

Table II shows performance results of five texture images. Similar to table I, the final rank also has been recorded. The RFS (SIFT) method has the best performance (ranks the first) and CNN-dictionary has second better performance (ranks the second) compared to others. In conclusion, we can see from table I and II that RFS (SIFT) has the best performance compared to the different deep learning methods. In addition, RFS (LF-net) and RFS (r2d2) show better consistent performance for texture and object images.

*D. Discussion*

Table I and II demonstrate the effect of using point pattern features within RFS for defect detection. Most of the current deep learning methods have structured data as input and produce structured feature during representation learning. Point pattern features detection methods provide sparse features (set of features) but these features are usually converted to structured data, such as using Bag of visual world (BoVW) [52]. Random finite set framework provides an elegant way to model feature set density that takes into account both cardinality and feature density. The main assumption here is that the defected samples will have more/less number of feature points compared to normal samples. The main reason why RFS (SIFT) works better or similar to the state-of-art methods is because RFS density considers both cardianlity and feature density information. The cardinality of these feature will have a distribution for normal samples and defect samples should have more/less number of features. It can be observed that using deep point feature detection, such as LF-net does not introduce any advantage compared to SIFT and the reason for this is that this model has not been trained on MVTec-AD dataset. However, most of these networks require ground-truth during the training and there is no ground-truth annotation for MVTec-AD dataset. Therefore, these deep learning methods perform poorly for feature extraction but still generate comparative results when combined with RFS-based defect detection.

TABLE II: Performance results of Poisson RFS-based defect detection on MVTec-AD dataset and other . The ratio of corrected classified normal samples (top row) and ratio of corrected classified abnormal samples (bottom row) are given. The best means are in **bold**.

| Category | RFS (SIFT) | RFS (Harris) | RFS (D2-net) | RFS (D2-net2) | RFS (LF-net) | RFS (r2d2) | AE (SSIM) | AE (L2) | AnoGAN | CNN | Texture inspection |
|---|---|---|---|---|---|---|---|---|---|---|---|
| Carpet | 0.71 | 0.28 | 0.96 | - | 0.42 | 0.42 | 0.43 | 0.57 | 0.82 | 0.89 | 0.57 |
|  | 0.70 | 0.83 | 0.08 | - | 0.76 | 0.79 | 0.90 | 0.42 | 0.16 | 0.36 | 0.61 |
| Mean | **0.70** | 0.55 | 0.52 | - | 0.59 | 0.61 | 0.66 | 0.49 | 0.49 | 0.62 | 0.59 |
| Rank | 1 | 6 | 7 | - | 5 | 4 | 2 | 8 | 8 | 3 | 5 |
| Grid | 0.52 | 0.52 | 0.66 | - | 0.00 | 0.80 | 0.38 | 0.57 | 0.90 | 0.57 | 1.00 |
|  | 0.86 | 0.52 | 0.56 | - | 1.00 | 0.36 | 1.00 | 0.98 | 0.12 | 0.33 | 0.05 |
| Mean | 0.69 | 0.52 | 0.61 | - | 0.50 | 0.58 | 0.69 | **0.77** | 0.51 | 0.45 | 0.52 |
| Rank | 2 | 5 | 3 | - | 7 | 4 | 2 | 1 | 6 | 8 | 5 |
| Leather | 0.93 | 0.34 | - | - | 0.00 | 0.34 | 0.00 | 0.06 | 0.91 | 0.63 | 0.00 |
|  | 0.40 | 0.73 | - | - | 1.00 | 0.78 | 0.92 | 0.82 | 0.12 | 0.71 | 0.99 |
| Mean | 0.66 | 0.54 | - | - | 0.50 | 0.56 | 0.46 | 0.44 | 0.51 | **0.67** | 0.49 |
| Rank | 2 | 4 | - | - | 6 | 3 | 8 | 9 | 5 | 1 | 7 |
| Wood | 1.00 | 1.00 | - | - | 0.94 | 0.94 | 0.84 | 1.00 | 0.89 | 0.79 | 0.42 |
|  | 0.78 | 0.31 | - | - | 0.56 | 0.30 | 0.82 | 0.47 | 0.47 | 0.88 | 1.00 |
| Mean | **0.89** | 0.65 | - | - | 0.75 | 0.62 | 0.83 | 0.73 | 0.68 | 0.83 | 0.71 |
| Rank | 1 | 7 | - | - | 3 | 8 | 2 | 4 | 6 | 2 | 5 |
| Tile | 0.42 | 0.24 | 1.00 | 1.00 | 0.78 | 0.33 | 1.00 | 1.00 | 0.97 | 0.97 | 1.00 |
|  | 0.70 | 0.86 | 0.00 | 0.01 | 0.55 | 0.80 | 0.04 | 0.54 | 0.05 | 0.44 | 0.43 |
| Mean | 0.56 | 0.55 | 0.50 | 0.50 | 0.67 | 0.57 | 0.52 | **0.77** | 0.51 | 0.70 | 0.71 |
| Rank | 5 | 6 | 9 | 9 | 3 | 4 | 7 | 1 | 8 | 2 | 2 |
| Avg.Rank | 2.2 | 5.6 | 6.3 | 9 | 4.8 | 4.6 | 4.2 | 4.6 | 6.6 | 3.2 | 4.8 |
| Rank | **1** | 6 | 7 | 9 | 5 | 4 | 3 | 4 | 8 | 2 | 5 |

## V. CONCLUSION

In this paper, an evaluation of different point pattern feature detection and description within Random finite set framework for defect detection has been presented. Random finite set framework has been used to model the cardinality and density of these features as a sophisticated way to build a model that best fits the normal samples by maximizing the log-likelihood. Different deep interest keypoint detection and descriptors have been used in this paper and the focus was on the data driven approaches (deep learning methods). The experimental results show that using keypoint feature extraction (especially SIFT) within RFS framework for defect detection has promising performance.